\documentclass[10pt,twocolumn,letterpaper]{article}

\usepackage{wacv}              %

\usepackage{graphicx}
\usepackage{amsmath}
\usepackage{amssymb}
\usepackage{booktabs}
\usepackage{multirow}

\usepackage[pagebackref,breaklinks,colorlinks]{hyperref}

\usepackage[capitalize]{cleveref}
\crefname{section}{Sec.}{Secs.}
\Crefname{section}{Section}{Sections}
\Crefname{table}{Table}{Tables}
\crefname{table}{Tab.}{Tabs.}

\def\wacvPaperID{1697} %
\def\confName{WACV}
\def\confYear{2025}

\begin{document}

\title{MemFusionMap: Working Memory Fusion for Online\\Vectorized HD Map Construction}

\author{Jingyu Song$\phantom{}^{1}$\thanks{Work done during an internship at NVIDIA. Corresponding at: \texttt{jingyuso@umich.edu}} \hspace{15pt}
Xudong Chen$\phantom{}^{2}$\hspace{15pt}
Liupei Lu$\phantom{}^{2}$\hspace{15pt}
Jie Li$\phantom{}^{2}$\hspace{15pt}
Katherine A. Skinner$\phantom{}^{1}$\vspace{8pt}\\
$\phantom{}^1$University of Michigan\hspace{5pt}
$\phantom{}^2$NVIDIA\hspace{5pt}
}

\maketitle

\begin{abstract}
    
   High-definition (HD) maps provide environmental information for autonomous driving systems and are essential for safe planning. While existing methods with single-frame input achieve impressive performance for online vectorized HD map construction, they still struggle with complex scenarios and occlusions. We propose MemFusionMap, a novel temporal fusion model with enhanced temporal reasoning capabilities for online HD map construction.
   Specifically, we contribute a working memory fusion module that improves the model's memory capacity to reason across a history of frames. We also design a novel temporal overlap heatmap to explicitly inform the model about the temporal overlap information and vehicle trajectory in the Bird's Eye View space. By integrating these two designs, MemFusionMap significantly outperforms existing methods while also maintaining a versatile design for scalability. We conduct extensive evaluation on open-source benchmarks and demonstrate a maximum improvement of 5.4\% in mAP over state-of-the-art methods.
   The project page for MemFusionMap is \url{https://song-jingyu.github.io/MemFusionMap}
\end{abstract}

\section{Introduction}
\label{sec:intro}

With recent advances in leveraging Bird's Eye View (BEV) representations, perception capabilities of Autonomous Vehicles (AVs) have been largely enhanced~\cite{li2022bevformer, philion2020lss, liu2023bevfusion, song2024lirafusion, zhao2024crkd}. Among all the BEV perception tasks, constructing High-Definition (HD) maps in an online manner has received increasing attention from the AV research community~\cite{liu2023vectormapnet, li2022hdmapnet, liao2023maptr, yuan2024streammapnet}. Specifically, online HD map construction aims to use onboard vehicle sensor data to reconstruct road elements such as lanes, pedestrian crossings and road boundaries on-the-fly. The reconstructed HD maps are essential for downstream planning tasks such as prediction and planning~\cite{gu2024producing}. This task is of great significance in advancing large-scale deployment of AVs as it avoids the need for a sophisticated and labor-intensive offline HD map generation process~\cite{liu2023vectormapnet}.

Existing works have demonstrated the effectiveness of formulating online HD map construction in an end-to-end framework~\cite{liu2023vectormapnet, liu2024mapqr, Qiao_2023_bemapnet, yuan2024streammapnet}. Though significant advances have been made in this field, many methods fall short of leveraging temporal information, which has been shown to be very beneficial, especially during instances of occlusion~\cite{yuan2024streammapnet, wang2024sqdmapnet, chen2024maptracker}.
StreamMapNet~\cite{yuan2024streammapnet} is the first end-to-end vectorized HD map construction work leveraging temporal cues. By maintaining a recurrent BEV memory feature map and propagating map queries, StreamMapNet conducts temporal fusion effectively and significantly outperforms previous non-temporal baselines~\cite{liu2023vectormapnet,liao2023maptr}, demonstrating the importance of incorporating temporal cues for online HD map construction. 
The success of StreamMapNet~\cite{yuan2024streammapnet} has inspired follow-up works to further improve it by incorporating auxiliary learning tasks or by integrating ideas from the tracking literature~\cite{wang2024sqdmapnet, chen2024maptracker}.

\begin{figure}[t]
    \centering
    \includegraphics[width=1.0\linewidth]{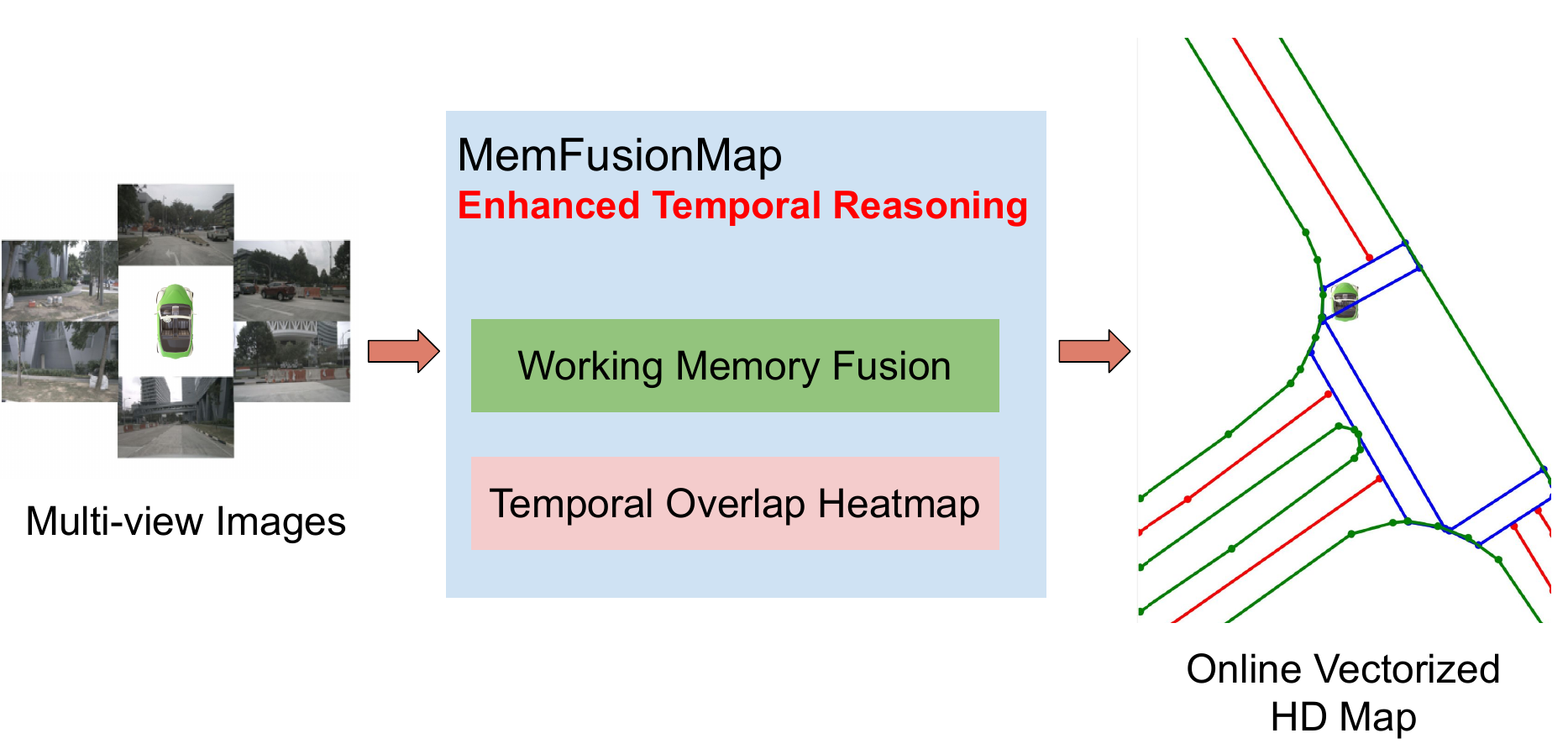}
    \caption{We propose MemFusionMap to improve the temporal reasoning capability for online vectorized HD map construction. MemFusionMap takes in multi-view images and leverages the proposed working memory fusion module and temporal overlap heatmap to output a vectorized HD map online.}
    \label{fig:pitch}
    \vspace{-6mm}
\end{figure}

Still, we highlight the challenges of directly accumulating all the temporal features into a single memory feature map, as proposed in~\cite{yuan2024streammapnet}. The network can struggle to reason about the entire history due to limited memory capacity in complex road environments since the network only has access to the latest memory feature and current BEV feature. Furthermore, this design can struggle with occlusion. A typical example is when a sudden occlusion occurs due to a moving truck near the ego-vehicle. This will trigger a bad update to the memory feature due to challenging 2D-to-BEV projection, and thus will affect all the future predictions. A similar issue is also noted in the video object segmentation field~\cite{cheng2022xmem, Zhou_2024_rmem}. We share the same insight that maintaining working memory features from a subset of historical frames is helpful, which is also pointed out by~\cite{chen2024maptracker}.
Lastly, we argue the importance of maintaining explicit temporal overlap information, as the model can leverage this input to reason about temporal information more effectively.

Given these insights, we propose MemFusionMap (\cref{fig:pitch}), a novel memory fusion framework designed for online vectorized HD Map construction. We design a working memory buffer to maintain working memory features in a fixed-lag manner. The working memory features are propagated recursively to align with the current field-of-view of the ego-vehicle. The fixed-lag design ensures bounded memory usage to fit real-world deployment needs.
In addition, we propose a novel use of a temporal overlap heatmap, represented as a BEV image with a single channel. The heatmap score indicates the number of times each grid cell falls within the field-of-view. We recurrently propagate the temporal overlap heatmap. To facilitate the model's temporal reasoning capabilities, the temporal overlap heatmap is fused with the working memory features by a convolutional temporal fusion block, whose output is fed into a standard detector head to generate road element predictions.

To summarize, our main contributions are as follows:
\begin{itemize}
    \item We propose a simple yet effective model to fuse working memory features in BEV space for online vectorized HD map construction. MemFusionMap focuses on improving the network's temporal reasoning capability while also maintaining a versatile design for scalability and compatibility.
    \item We propose a novel design of maintaining a temporal overlap heatmap, providing a strong cue for the model to reason across a history of frames and also implicitly encoding valuable insights of the vehicle's trajectory.
    \item We conduct extensive evaluation on nuScenes~\cite{caesar2020nuscenes} and Argoverse2~\cite{wilson2023argoverse} to demonstrate the effectiveness of MemFusionMap. The proposed method significantly outperforms the state-of-the-art method~\cite{yuan2024streammapnet}, achieving a maximum improvement of $5.4\%$ in mAP.
    
\end{itemize}

\begin{figure*}[t!]
    \centering
    \includegraphics[width=1.0\linewidth]{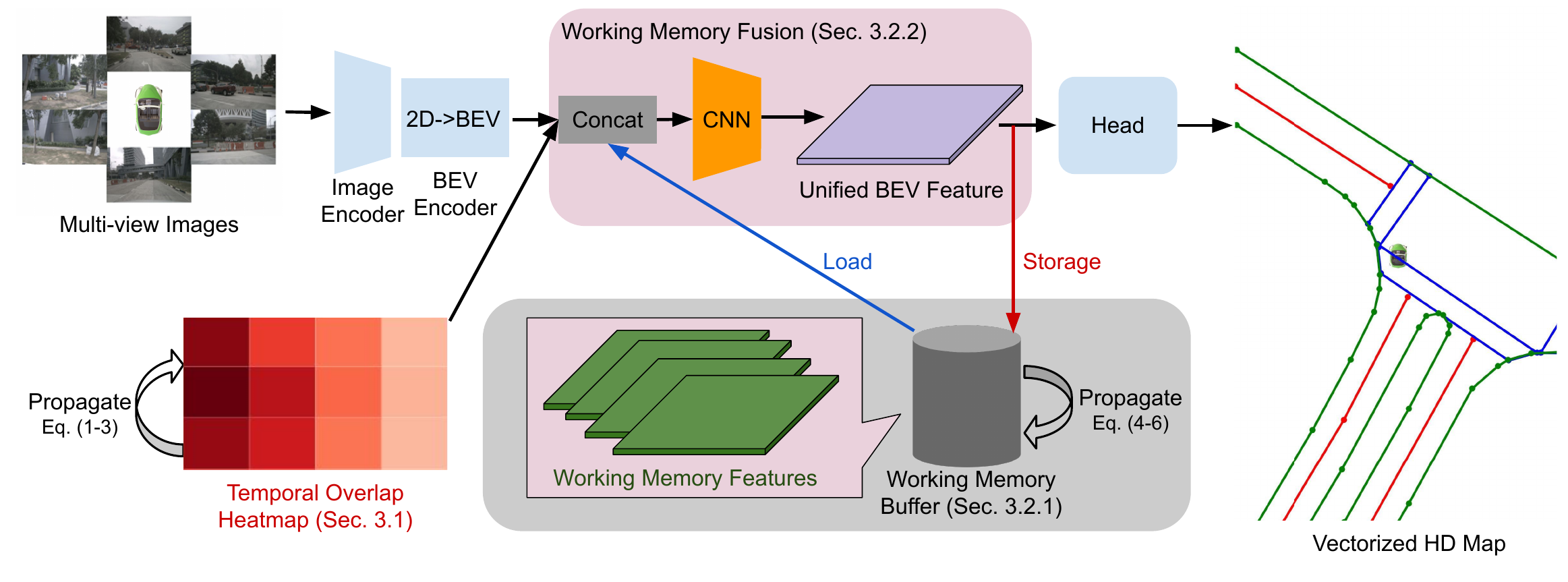}
    \caption{Overall architecture of MemFusionMap. MemFusionMap takes in multi-view images and extracts a BEV feature via an image encoder and a BEV feature encoder. The encoded BEV feature is fused with working memory features and the temporal overlap heatmap to obtain a unified BEV feature. The unified feature is fed into a decoder head to generate HD map prediction.}
    \label{fig:overall_architecture}
    \vspace{-5mm}
\end{figure*}

\section{Related Works}
\subsection{Online Vectorized HD Map Construction}
Constructing HD maps is one of the key tasks for AV perception. While traditional SLAM-based methods~\cite{zhang2014loam, shan2018lego} suffer from high cost during offline construction and maintenance, online HD map construction solutions have obtained increasing attention as they can effectively simplify the map construction process and handle map changes.
HDMapNet~\cite{li2022hdmapnet} marks a significant step of moving towards vectorized HD map representations, which are preferred by the downstream tasks over the rasterized map representations that are generated from previous segmentation methods~\cite{li2022bevformer, philion2020lss, zhou2022cross} or lane detection methods~\cite{feng2022rethinking_lane_detection, chen2022persformer, wang2023bevlanedet}. VectorMapNet~\cite{liu2023vectormapnet} proposes the first end-to-end vectorized HD map construction framework where a DETR~\cite{carion2020detr} based map decoder is used to generate the map elements' prediction with an auto-regressive transformer. Its performance is further improved by MapTR~\cite{liao2023maptr} and MapTRV2~\cite{liao2023maptrv2} by treating the map construction task as a point set prediction problem and adding auxiliary tasks and decoder design improvement.
Other works focus on different map element representations: BeMapNet~\cite{Qiao_2023_bemapnet} and PivotNet~\cite{Ding_2023_pivotnet} explore using Bézier curves and pivot-based points, respectively. Additionally, other methods investigate network design improvement and training pipeline enhancements~\cite{Liu_2024_mgmap, liu2024mapqr, xu2023insightmapper, chen2024polydiffuse}. More recently, StreamMapNet~\cite{yuan2024streammapnet} and its follow-up works~\cite{chen2024maptracker, wang2024sqdmapnet} demonstrate notable improvement by leveraging temporal fusion for this task, which will be discussed in \cref{sec:related_works_temporal_fusion}.

\subsection{Temporal Fusion for HD Map Construction}
\label{sec:related_works_temporal_fusion}
Fusing temporal information has proven effective in various works in domains like 3D object detection and scene completion~\cite{han2024videobev, wilson2022motionsc, Wang_2023_streampetr, park2022solofusion, li2022bevformer}. HDMapNet~\cite{li2022hdmapnet} is the first to demonstrate the benefits of temporal fusion in online HD map construction. It directly applies max pooling to compress temporal information into a BEV feature map, which is fed into the decoder. StreamMapNet~\cite{yuan2024streammapnet} develops the first temporal fusion pipeline in an end-to-end framework for this task. In addition to propagating and reusing queries in the decoder, it proposes a BEV temporal fusion paradigm leveraging the streaming strategy following~\cite{Wang_2023_streampetr, han2024videobev}. StreamMapNet~\cite{yuan2024streammapnet} compresses the history of information into a latent memory feature and propagates the latent memory feature recurrently. It applies a Gated Recurrent Unit (GRU)~\cite{chung2014GRU} to fuse the latent memory feature with the current BEV feature from the BEV feature encoder.

The success of StreamMapNet has motivated subsequent works such as SQD-MapNet~\cite{wang2024sqdmapnet} and MapTracker~\cite{chen2024maptracker}. Inspired by DN-DETR~\cite{Li_2022_DN-DETR}, SQD-MapNet~\cite{wang2024sqdmapnet} designs a stream query denoising approach to further improve over StreamMapNet. However, SQD-MapNet is only evaluated on the original splits of nuScenes~\cite{caesar2020nuscenes} and Argoverse2~\cite{wilson2023argoverse} that are prone to overfitting, contradicting the main motivation of online HD map construction. MapTracker~\cite{chen2024maptracker} explores leveraging additional ground truth of tracked road elements to enhance the query propagation paradigm. MemFusionMap shares the same intuition as MapTracker~\cite{chen2024maptracker} that the redundancy of maintaining a subset of past frames can improve the model's robustness. However, we design MemFusionMap with a strong emphasis on scalability and versatility for practical application. Notably, while MapTracker fuses the last 20 frames of memory latents selected according to the distance strides, our proposed metohd, MemFusionMap, designs a simple yet effective memory fusion module, which only uses 4 past frames to avoid computational overhead and to achieve better training efficiency and runtime. MemFusionMap also covers the perception range of $100\times50\,m$, which has more practical value for real-world AV deployment, while MapTracker~\cite{chen2024maptracker} omits this extended perception range. Furthermore, SQD-MapNet~\cite{wang2024sqdmapnet} and MapTracker~\cite{chen2024maptracker} require implementing additional task and loss modules, which can cause extra overhead when integrating into existing multi-task AV perception pipelines. MapTracker also requires additional post-processing to generate the tracks of map elements, which can be time-consuming and potentially incompatible with large-scale databases that use an auto labeling process that cannot guarantee quality assured labels for all frames. In contrast, MemFusionMap maintains a compact design with better versatility that can be directly applied to existing pipelines by switching the BEV temporal fusion module. Lastly, MemFusionMap also proposes a novel method for maintaining and integrating a temporal overlap heatmap, which is demonstrated to significantly improve the model's temporal reasoning capability.

\section{MemFusionMap}

\label{sec:method_overall_architecture}
The overall framework of MemFusionMap is shown in \cref{fig:overall_architecture}. The pipeline of MemFusionMap is inspired by StreamMapNet~\cite{yuan2024streammapnet}. However, MemFusionMap proposes a novel structure for the temporal BEV fusion module. MemFusionMap firstly uses a shared ResNet-50~\cite{He_2016_resnet} image encoder to extract image features. A BEV feature encoder further projects the image features in 2D to BEV space. The projected BEV feature is denoted as $\mathcal{F}_\mathrm{BEV} \in \mathbb{R}^{C\times H\times W}$, where $C$ is the BEV feature dimension, and $H$ and $W$ represent the spatial dimensions of the BEV feature. The BEV feature is fed into the memory fusion block, which will be introduced in the following sections. The output of the memory fusion block is a unified BEV feature map containing temporal cues, denoted as $\tilde{\mathcal{F}}_\mathrm{BEV}\in \mathbb{R}^{C\times H\times W}$.
Following~\cite{yuan2024streammapnet}, we use a DETR~\cite{carion2020detr} based transformer decoder with a multi-point attention mechanism for extended perception range. We also maintain a memory buffer to properly propagate and reuse history map queries.

The following sub-sections discuss the key contributions of the proposed architecture in more detail.
The core contribution of MemFusionMap is its temporal BEV fusion design, which consists of a temporal overlap heatmap (\cref{sec:method_overlap_heatmap}) and the working memory fusion module (\cref{sec:method_memory_fusion}).

\subsection{Temporal Overlap Heatmap}
\label{sec:method_overlap_heatmap}

\begin{figure}[t]
    \centering
    \includegraphics[width=0.97\linewidth]{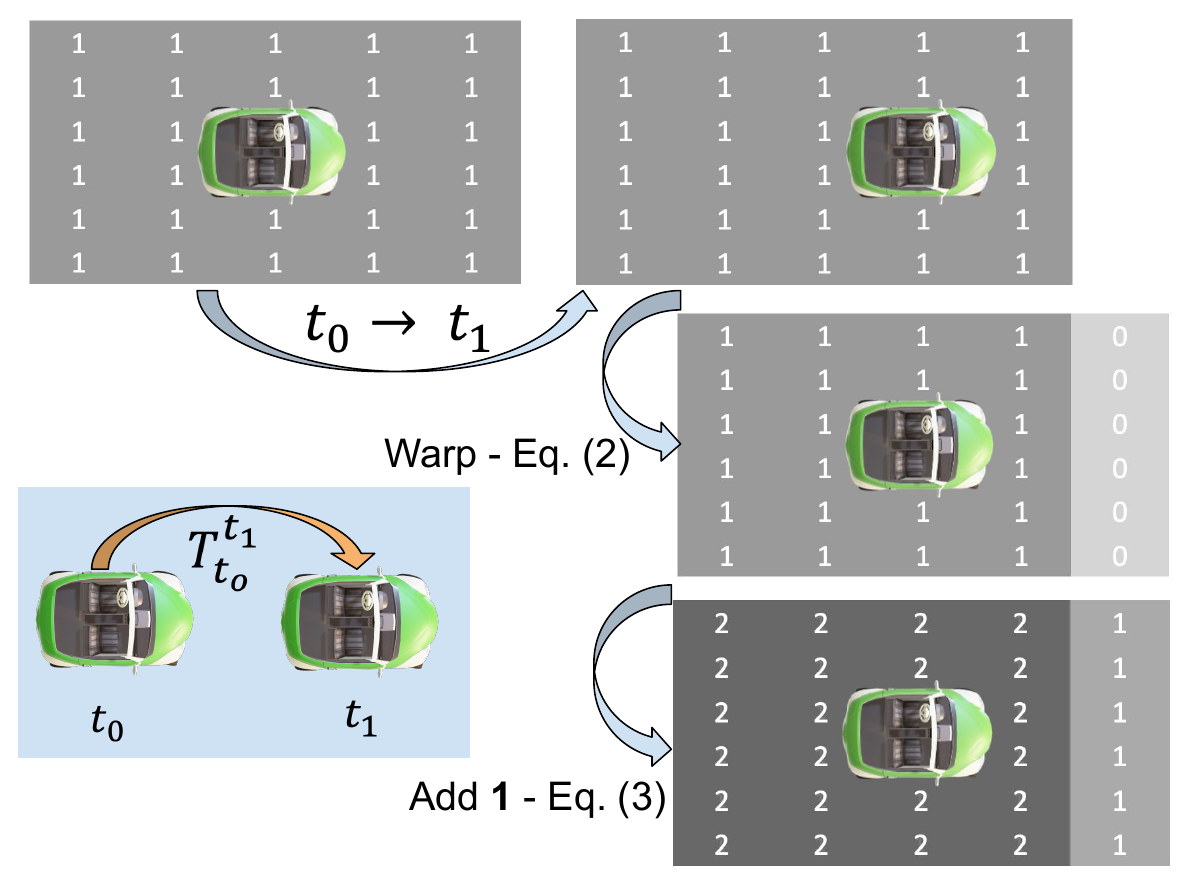}
    \caption{Example propagation process of the temporal overlap heatmap from $t_0$ to $t_1$.}
    \label{fig:overlap_propagate}
    \vspace{-5mm}
\end{figure}

We propose a novel paradigm to maintain a temporal overlap heatmap to explicitly inform the model of the temporal overlap information. %
Figure \ref{fig:overlap_propagate} illustrates the process of maintaining the heatmap. Starting from the first frame of each clip, we initialize the temporal overlap heatmap $\mathbf{H}_{t_0}$ as follows:

\begin{equation}
\label{eq: warp}
\mathbf{H}_{t_0} = \mathbf{1}^{1 \times H \times W}.
\end{equation}
Specifically, $\mathbf{H}_{t_0}$ has the same spatial size as $\mathcal{F}_\mathrm{BEV}$ and is single-channel. Each pixel of $\mathbf{H}$ indicates the temporal overlap score of this cell in the corresponding BEV space. $\mathbf{H}_{t_0}$ is initialized with all 1s, meaning the whole BEV has not been seen before. Then, as the second frame ($t_1$) comes in, we first propagate the temporal overlap heatmap to the new vehicle position as follows:
\begin{equation}
    \hat{\mathbf H}_{t_1} = \mathrm{Warp}(\mathbf{H}_{t_0}, T_{t_0}^{t_1}),
\end{equation}
where $\hat{\mathbf H}_{t_1}$ is the propagated and warped overlap heatmap from $t_0$ to $t_1$ according to the egomotion $T_{t_0}^{t_1}$. The warping is implemented using the grid sample function in PyTorch with zero padding mode for the out-of-bound grids. We then obtain the final heatmap for $t_1$ as:
 \begin{equation}
    \mathbf H_{t_1} = \hat{\mathbf H}_{t_1} + \mathbf{1}^{1 \times H \times W}.
\end{equation}
Following this strategy, we keep updating the temporal overlap heatmap recurrently. At each timestamp, the overlap heatmap is fed into the memory fusion block, which will be discussed in \cref{sec:method_memory_fusion}.

One main benefit of maintaining this temporal overlap heatmap is that it provides informative insights into the regions with heavy overlaps and those without. With this information, the model can more adaptively reason across the current BEV feature and the historical features. Specifically, in the overlap heatmap $\mathbf{H}_t$, a smaller value indicates that a region is less temporally overlapped, meaning more information from the current BEV feature should be trusted. Conversely, a larger value suggests a region with greater overlap, implying that information from the memory latents could be more valuable. Furthermore, the overlap heatmap naturally encodes the vehicle trajectory, providing the model with additional knowledge about vehicle motion to enable more adaptive temporal reasoning.

\begin{figure}[t]
    \centering
    \includegraphics[width=1.0\linewidth]{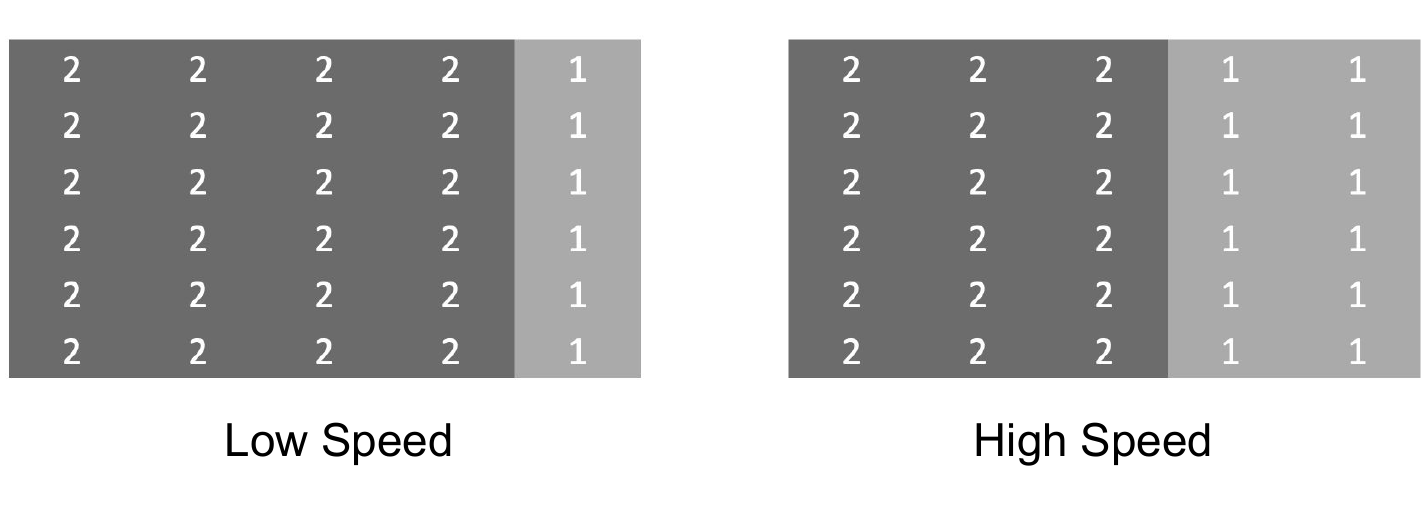}
    \caption{Example temporal overlap heatmap resulted from different speeds. The comparison demonstrates that the overlap heatmap encodes vehicle moving speed.}
    \label{fig:overlap_benefit1}
    \vspace{-4mm}
\end{figure}

\begin{figure}[h]
    \centering
    \includegraphics[width=1.0\linewidth]{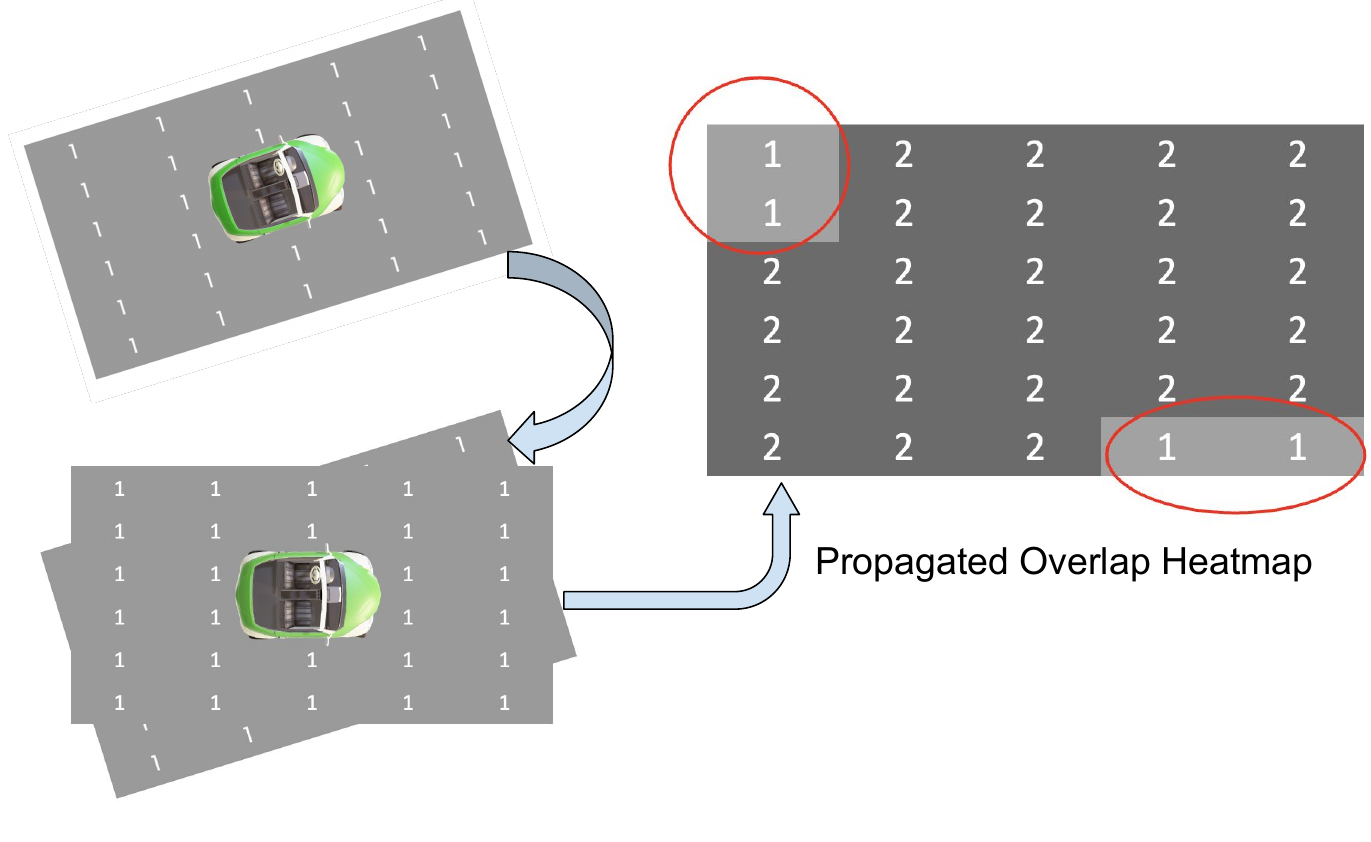}
    \caption{Example temporal overlap heatmap resulted from a vehicle turning. We highlight the two noticeable edges at the corner. This pattern encodes insights into the vehicle's trajectory.}
    \label{fig:overlap_benefit2}
    \vspace{-6mm}
\end{figure}

We demonstrate this benefit with two visual examples. As shown in \cref{fig:overlap_benefit1}, we present two example overlap heatmaps resulting from different speeds. The heatmap on the right exhibits a larger area with a value of $\mathbf{1}$, indicating that a larger area is newly seen, which is a result of higher-speed egomotion. 
We present another example of a vehicle turning in \cref{fig:overlap_benefit2}. 
The propagated overlap heatmap reveals two noticeable edges at the corner. These low-level feature patterns can be easily detected by a convolutional neural network, providing critical insights into the vehicle's trajectory.

\subsection{Memory Fusion Block}
\label{sec:method_memory_fusion}
We design a memory fusion block to synthesize memory features with the temporal overlap heatmap. As discussed in \cref{sec:intro}, we highlight the importance of increasing the memory capacity for better temporal reasoning capability and robustness against occlusion. We aggregate a fixed number of BEV features from historical frames to fuse with the current BEV feature. We follow~\cite{cheng2022xmem} to use the terminology used in cognitive science~\cite{doherty2019dual, cowan2010magical, xie2022working} and refer to the subset of historical frames as working memory features. As shown in \cref{fig:overall_architecture}, we develop a working memory buffer to properly handle storage and loading of working memory features. We also design a memory fusion block to temporally reason across working memory features, aided by the temporal overlap heatmap.
\subsubsection{Working Memory Buffer}
We design a working memory buffer to manage the memory features and update them recurrently. In the streaming training schedule, at time $t$, before memory fusion, the working memory feature $\mathcal{F}_\mathrm{WM}^t$ is defined as follows:
\begin{equation}
    \mathcal{F}_\mathrm{WM}^t = \{\tilde{\mathcal{F}}_{\text{BEV}}^{t - T_{\text{WM}}}, \tilde{\mathcal{F}}_{\text{BEV}}^{t - T_{\text{WM}} + 1}, \ldots, \tilde{\mathcal{F}}_{\text{BEV}}^{t - 2}, \tilde{\mathcal{F}}_{\text{BEV}}^{t-1}\},
\end{equation}
where $\tilde{\mathcal{F}}_{\text{BEV}}^{t}$ is the unified BEV feature after memory fusion at time $t$, and $T_\text{WM}$ is the working memory capacity. Progressing towards time $t+1$, we first add the new feature and drop the oldest feature:
\begin{equation}
    \mathcal{F}_\mathrm{WM}^{t+1} = \{\tilde{\mathcal{F}}_{\text{BEV}}^{t - T_{\text{WM}} + 1}, \tilde{\mathcal{F}}_{\text{BEV}}^{t - T_{\text{WM}} + 2}, \ldots, \tilde{\mathcal{F}}_{\text{BEV}}^{t - 1}, \tilde{\mathcal{F}}_{\text{BEV}}^{t}\}.
\end{equation}
We also warp all the working memory features to align with $\mathcal{F}_{\text{BEV}}^{t+1}$:
\begin{equation}
    \mathcal{F}_\mathrm{WM}^{t+1} = \mathrm{Warp}(\mathcal{F}_\mathrm{WM}^{t+1}, T_{t}^{t+1}),
\end{equation}
where $T_{t}^{t+1}$ is the transformation from $t$ to $t+1$. Additionally, at $t_0$, the beginning of a sequence, we initialize $\mathcal{F}_\mathrm{WM}^{t_0}$ by repeating $\mathcal{F}_{\text{BEV}}^{t_0}$ for $T_{\text{WM}}$ times. After the memory fusion at $t_0$, we replace $\mathcal{F}_{\text{BEV}}^{t_0}$ with $\tilde{\mathcal{F}}_{\text{BEV}}^{t_0}$ and warp them to form $\mathcal{F}_{\text{BEV}}^{t_1}$. The following process of updating working memory follows the strategy mentioned above. Meanwhile, the working memory buffer also serves to store the temporal overlap heatmap. The propagation process for the heatmap is discussed in \cref{sec:method_overlap_heatmap}.

\subsubsection{Working Memory Fusion}
We propose a working memory fusion module to obtain a unified BEV feature $\tilde{\mathcal{F}}_{\text{BEV}}$ fused with temporal cues. The goal of the memory fusion block is to effectively utilize the temporal information in working memory features, aided by the temporal overlap heatmap, and to fuse them with the current BEV feature to obtain a unified BEV feature with rich temporal information.

Specifically, at time $t$, the input to this module consists of the working memory features $\mathcal{F}_\mathrm{WM}^{t} \in \mathbb{R}^{C_{\text{WM}}\times H\times W}$, the temporal overlap heatmap $\mathbf{H}_t \in \mathbb{R}^{1\times H\times W}$, and the BEV feature $\mathcal{F}_\text{BEV}^t \in \mathbb{R}^{C\times H\times W}$ obtained from the BEV feature encoder. Here $C$ is the feature channel dimension of the BEV feature encoder, $C_{\text{WM}} = T_{\text{WM}}\times C$, and $H$ and $W$ represent the spatial size of the BEV feature. The output of the working memory fusion module is the unified feature $\tilde{\mathcal{F}}_\text{BEV}^t \in \mathbb{R}^{C\times H\times W}$. In the working memory fusion module, the first step is to extract low-level features from the temporal overlap heatmap:
\begin{equation}
    \tilde{\mathbf{H}}_t = \text{sigmoid}(\text{Conv}_{\mathbf{H}}(\mathbf{H}_t)) \in \mathbb{R}^{C_{\mathbf{H}} \times H\times W},
\end{equation}
where $\text{Conv}_{\mathbf{H}}$ is a convolution block composed of 3 convolution layers with ReLU, and $C_{\mathbf{H}}$ is the feature dimension of the learned temporal overlap heatmap feature. We apply a sigmoid function on the output temporal overlap feature to bound its value. Subsequently, we use a convolutional memory fusion block to obtain the unified BEV feature $\tilde{\mathcal{F}}_\text{BEV}^t$ following:
\begin{equation}
    \tilde{\mathcal{F}}_\text{BEV}^t = \text{LayerNorm}(\text{Conv}_{\text{Mem}}(\text{Concat}(\mathcal{F}_\mathrm{WM}^{t}, \tilde{\mathbf{H}}_t, \mathcal{F}_\text{BEV}^t))),
\end{equation}
where $\text{Conv}_{\text{Mem}}$ is composed of 3 convolution layers with ReLU. We follow~\cite{yuan2024streammapnet} to apply layer normalization to improve training stability.

When designing the working memory fusion module, we keep the HD map construction task in mind. We observe that a unique feature for this task is that map elements, such as lane markings, can have elongated shapes, which inspires us to enhance the receptive field of the memory fusion module to let temporal features interact more effectively both temporally and spatially. Therefore, we apply dilated (atrous) convolution~\cite{chen2017dilation} in $\text{Conv}_{\text{Mem}}$ to further improve the working memory fusion module.

\subsection{Training Loss}
\label{sec:method_training_pipeline}
During training, the overall map loss $\mathcal{L}_{map}$ is defined as in~\cite{yuan2024streammapnet}:
\begin{equation}
    \mathcal{L}_{map} = \lambda_1 \mathcal{L}_{Focal} + \lambda_2 \mathcal{L}_{line} + \lambda_3 \mathcal{L}_{trans},
\end{equation}
where $\mathcal{L}_{Focal}$ is the classification matching cost, $\mathcal{L}_{line}$ is the polyline-wise matching cost, $\mathcal{L}_{trans}$ is the auxiliary transformation loss for temporal query propagation, and $\lambda_1$-$\lambda_3$ are pre-defined loss weights.
We refer to~\cite{yuan2024streammapnet} for more details about the loss function.

\section{Experiments}

\begin{table*}[t]
  \centering
  \resizebox{0.96\textwidth}{!}{
    \begin{tabular}{c|lccc|cccc}
\hline
Range                              & Method              & Backbone & Image Size      & Epoch & AP$_{ped}$    & AP$_{div}$    & AP$_{bound}$  & mAP           \\ \hline
\multirow{4}{*}{$60\times 30\,m$}  & VectorMapNet$^\dagger$~\cite{liu2023vectormapnet}        & R50      & $256\times 480$ & 120   & 15.8          & 17.0          & 21.2          & 18.0          \\
                                   & MapTR$^\dagger$~\cite{liao2023maptr}               & R50      & $480\times 800$ & 24    & 6.4           & 20.7          & 35.5          & 20.9          \\
                                   & StreamMapNet~\cite{yuan2024streammapnet}        & R50      & $480\times 800$ & 24    & 30.7          & 29.9          & 41.8          & 34.1          \\
                                   & MemFusionMap (ours) & R50      & $480\times 800$ & 24    & \textbf{38.3} & \textbf{32.1} & \textbf{43.6} & \textbf{38.0} \\ \hline
\multirow{4}{*}{$100\times 50\,m$} & VectorMapNet$^\dagger$~\cite{liu2023vectormapnet}        & R50      & $256\times 480$ & 120   & 12.0          & 8.1           & 6.3           & 8.8           \\
                                   & MapTR$^\dagger$~\cite{liao2023maptr}               & R50      & $480\times 800$ & 24    & 8.3           & 16.0          & 20.0          & 14.8          \\
                                   & StreamMapNet~\cite{yuan2024streammapnet}        & R50      & $480\times 800$ & 24    & 25.4          & 19            & 22.2          & 22.2          \\
                                   & MemFusionMap (ours) & R50      & $480\times 800$ & 24    & \textbf{30.2} & \textbf{22.6} & \textbf{29.9} & \textbf{27.6} \\ \hline
\end{tabular}
  }
  \caption{Performance comparison with baseline methods on the new nuScenes split at both $30\,m$ and $50\,m$ perception ranges. $^\dagger$: results reported in StreamMapNet~\cite{yuan2024streammapnet} paper. Abbreviations: pedestrian crossing (ped), lane divider (div), road boundariy (bound). For StreamMapNet, we report our reproduced results, which match the original paper.}
  \label{tab:nusc_new}
  \vspace{-6mm}
\end{table*}

\subsection{Datasets and Splits}
We evaluate our methods on the open-source nuScenes~\cite{caesar2020nuscenes} and Argoverse2~\cite{wilson2023argoverse} benchmarks. The nuScenes benchmark has an annotation keyframe rate of 2Hz, with 6 cameras synchronized. Argoverse2 is annotated with 10Hz with 7 ring cameras and 2 stereo cameras. We follow~\cite{yuan2024streammapnet} to unify the frame rate of the Argoverse2 dataset to 2Hz, and use all 6 cameras of nuScenes and the 7 ring cameras of Argoverse2 as the input to the model.

As noted by~\cite{Roddick_2020_nusc_new_split, yuan2024streammapnet}, the default splits in both nuScenes and Argoverse2 can be prone to overfitting as there is a large portion of overlap between training and validation sets. We highlight that the ultimate goal of the online HD map construction task is to develop a generalizable model that can adapt to unseen environments with robustness to potential map changes, instead of simply memorizing the training set. Therefore, we focus on evaluating MemFusionMap using the split proposed in~\cite{Roddick_2020_nusc_new_split} for nuScenes and the split proposed in~\cite{yuan2024streammapnet} for Argoverse2. We refer to these non-overlapped splits as new splits and the default splits as original splits in the following sub-sections.

\subsection{Evaluation Metrics}
We follow~\cite{yuan2024streammapnet,wang2024sqdmapnet} to consider both a small perceptual range ($60\times 30m$) and a large perceptual range ($100\times 50m$). We mainly use the geofenced new splits proposed by~\cite{Roddick_2020_nusc_new_split, yuan2024streammapnet}. We use Average Precision (AP) as the evaluation metric. We calculate AP using the distance threshold of $\{0.5m, 1.0m, 1.5m\}$ for the small perceptual range and  $\{1.0m, 1.5m, 2.0m\}$ for the large perceptual range. Following common practices in existing works, we evaluate three types of map elements: pedestrian crossings, lane dividers, and road boundaries. We calculate the mean AP (mAP) score of the three types of map elements.

\begin{table*}[t]
  \centering
  \resizebox{0.96\textwidth}{!}{
    \begin{tabular}{c|lccc|cccc}
\hline
Range                              & Method              & Backbone & Image Size      & Epoch & AP$_{ped}$    & AP$_{div}$    & AP$_{bound}$  & mAP           \\ \hline
\multirow{4}{*}{$60\times 30\,m$}  & VectorMapNet$^\dagger$~\cite{liu2023vectormapnet}        & R50      & $384\times 384$ & 120   & 35.6          & 34.9          & 37.8          & 36.1          \\
                                   & MapTR$^\dagger$~\cite{liao2023maptr}               & R50      & $608\times 608$ & 30    & 48.1          & 50.4          & 55            & 51.1          \\
                                   & StreamMapNet~\cite{yuan2024streammapnet}        & R50      & $608\times 608$ & 30    & 57.5          & 55.9          & 61.5          & 58.3          \\
                                   & MemFusionMap (ours) & R50      & $608\times 608$ & 30    & \textbf{59.3} & \textbf{57.2} & \textbf{65.1} & \textbf{60.6} \\ \hline
\multirow{4}{*}{$100\times 50\,m$} & VectorMapNet$^\dagger$~\cite{liu2023vectormapnet}        & R50      & $384\times 384$ & 120   & 32.4          & 20.6          & 24.3          & 25.7          \\
                                   & MapTR$^\dagger$~\cite{liao2023maptr}               & R50      & $608\times 608$ & 30    & 46.3          & 36.3          & 38            & 40.2          \\
                                   & StreamMapNet~\cite{yuan2024streammapnet}        & R50      & $608\times 608$ & 30    & 60.2          & 45.5          & 49            & 51.5          \\
                                   & MemFusionMap (ours) & R50      & $608\times 608$ & 30    & \textbf{64.1} & \textbf{47.3} & \textbf{51.5} & \textbf{54.3} \\ \hline
\end{tabular}
  }
  \caption{Performance comparison with baseline methods on the new Argoverse2 split at both $30\,m$ and $50\,m$ perception ranges. $^\dagger$: results reported in StreamMapNet~\cite{yuan2024streammapnet} paper. Abbreviations: pedestrian crossing (ped), lane divider (div), road boundariy (bound). For StreamMapNet, we report our reproduced results, which match the original paper.}
  \label{tab:av2_new}
  \vspace{-5mm}
\end{table*}

\subsection{Implementation Details}
We build MemFusionMap upon the codebase of StreamMapNet~\cite{yuan2024streammapnet}. We use ResNet-50~\cite{He_2016_resnet} as the image feature encoder. We use BEVFormer~\cite{li2022bevformer} as the BEV feature encoder to project features from 2D to BEV space. $C$ is set as 256 across all experiments. We set $H$ as 50 and $W$ as 100. In the memory fusion module, we set $T_{\text{WM}}$ as 4 and $C_{\mathbf{H}}$ as 32.  
The model is trained on 8 NVIDIA V100 GPUs with a total batch size of 32. The learning rate is set to $5\times10^{-4}$ and an AdamW~\cite{loshchilov2017adamw} optimizer is used. To train MemFusionMap, we apply the streaming training strategy~\cite{Wang_2023_streampetr, han2024videobev} to detach the memory tensor so gradients do not propagate back to previous frames. Following~\cite{park2022solofusion, yuan2024streammapnet}, we apply a two-stage training procedure, where the input of the first stage is single-frame. It is followed by a temporal training stage, where the input is consecutive sequences obtained from dividing the raw training sequence into 2 random sequences.
We include more implementation details in the supplementary material.

\subsection{Quantitative Comparison}
We compare MemFusionMap against state-of-the-art (SOTA) baselines across the nuScenes~\cite{caesar2020nuscenes} and Argoverse2~\cite{wilson2023argoverse} datasets with two perceptual ranges using the new splits~\cite{Roddick_2020_nusc_new_split, yuan2024streammapnet}. The baselines in the comprehensive comparison include StreamMapNet~\cite{yuan2024streammapnet} and single-frame baselines, including VectorMapNet~\cite{liu2023vectormapnet} and MapTR~\cite{liao2023maptr}. As to the follow-up works of StreamMapNet (i.e., SQD-MapNet~\cite{wang2024sqdmapnet} and MapTracker~\cite{chen2024maptracker}), we conduct separate comparisons. This is due to the lack of support for either the extended $100\times50\,m$ perception range in~\cite{chen2024maptracker}, which holds greater practical value for AV deployment, or the absence of open-source implementation and results evaluated using the new splits in~\cite{wang2024sqdmapnet}. We report the number of epochs to demonstrate the training efficiency and measure the runtime on a desktop equipped with an NVIDIA A6000 GPU.

We first show an overall comparison on the new split~\cite{Roddick_2020_nusc_new_split} of the nuScenes~\cite{caesar2020nuscenes} benchmark. At both ranges, MemFusionMap achieves significant improvement over baselines across all metrics, demonstrating the effectiveness of the proposed memory fusion architecture. Similarly, as shown in \cref{tab:av2_new}, MemFusionMap demonstrates superior performance over other methods across all categories on the Argoverse2 benchmark, validating the consistent improvement brought by MemFusionMap. We also highlight the greater improvement in the long perception range setting compared to the short range setting on both benchmarks, which is more applicable to real-world deployment. As MemFusionMap is built upon StreamMapNet and shares a similar training objective, we emphasize the advancement brought by MemFusionMap over StreamMapNet, which directly shows the improvement brought by the proposed working memory fusion framework over the previous GRU-based temporal fusion paradigm.

Furthermore, we show a comparison with open-source temporal baselines on mAP and runtime in \cref{tab:quantitative_maptracker}. Though MemFusionMap significantly outperforms StreamMapNet~\cite{yuan2024streammapnet} while maintaining a similarly fast runtime, it falls behind the recent SOTA work MapTracker~\cite{chen2024maptracker} in mAP. We recognize the contribution of MapTracker of formulating the HD map construction task as a tracking problem, and think merging MemFusionMap and MapTracker is an interesting future direction. However, it is worth mentioning that MapTracker is benefited from additional ground truth of tracked map elements and requires a time-consuming post-processing process. This may have scalability and compatibility issues with large-scale databases that use an auto labeling process that cannot guarantee all frames are quality assured, which could render a direct comparison on mAP inappropriate.
Additionally, as shown in \cref{tab:quantitative_maptracker}, MemFusionMap demonstrates strong performance without requiring a long training schedule, which is very valuable for integrating to large-scale databases. Compared with MapTracker~\cite{chen2024maptracker}, MemFusionMap also improves runtime significantly. Considering these factors, we still think MemFusionMap has unique benefits for the community.

\begin{table}[t]
\centering
\begin{tabular}{lc|cc}
\hline
Method              & Epoch & mAP  & FPS         \\ \hline
StreamMapNet~\cite{yuan2024streammapnet}        & 24    & 34.1 & \textbf{15.3}\\
MapTracker~\cite{chen2024maptracker}          & 72    & \textbf{40.3} & 11.6\\
MemFusionMap (ours) & 24    & 38.0 & 15.1\\ \hline
\end{tabular}
\caption{Quantitative comparison of training efficiency and runtime on the new nuScenes split at $30\,m$ range. MapTracker~\cite{chen2024maptracker} lacks support for the extended $50\,m$ range setting. SQD-MapNet~\cite{wang2024sqdmapnet} is not included as it is not open-source.}
\label{tab:quantitative_maptracker}
\vspace{-3mm}
\end{table}

We also show additional comparison on the original split (\cref{tab:quantitative_sqd}) for completeness since SQD-MapNet~\cite{wang2024sqdmapnet} does not report results on the new split and is not open-source. The evaluation is based on the $60\times30\,m$ range since SQD-MapNet uses a larger BEV spatial dimension for the extended range. Results in \cref{tab:quantitative_sqd} demonstrate that MemFusionMap still outperforms the baselines, though we note that this split is prone to overfitting.

\begin{table}[t]
\begin{tabular}{c|lc|c}
\hline
Dataset                     & Method              & \multicolumn{1}{l|}{Epoch} & \multicolumn{1}{l}{mAP} \\ \hline
\multirow{3}{*}{nuScenes}   & StreamMapNet~\cite{yuan2024streammapnet}        & $30^*$                         & 63.4                    \\
                            & SQD-MapNet~\cite{wang2024sqdmapnet}          & 24                         & 65                      \\
                            & MemFusionMap (ours) & $30^*$                         & \textbf{65.2}           \\ \hline
\multirow{3}{*}{Argoverse2} & StreamMapNet$^\dagger$~\cite{yuan2024streammapnet}        & 30                         & 61.5                    \\
                            & SQD-MapNet~\cite{wang2024sqdmapnet}          & 30                         & 63.3                    \\
                            & MemFusionMap (ours) & 30                         & \textbf{63.8}           \\ \hline
\end{tabular}
\caption{Performance comparison on the original split of nuScenes at $30\,m$ range. $^*$: we follow the practice of the StreamMapNet code repository to train 30 epochs on the original split. $^\dagger$: results reported in SQD-MapNet~\cite{wang2024sqdmapnet} paper.}
  \label{tab:quantitative_sqd}
  \vspace{-6mm}
\end{table}

\begin{table}[t]
\centering
\begin{tabular}{c|l|c}
\hline
Index & Method                       & mAP           \\ \hline
(a)   & StreamMapNet                 & 22.2          \\
(b)   & $-$ GRU BEV fusion           & 19.5          \\
(c)   & $+$ Working Memory Fusion    & 25.5          \\
(d)   & $+$ Dilated convolution      & 25.9          \\
(e)   & $+$ Temporal overlap heatmap & 27.6          \\ \hline
      & MemFusionMap                 & \textbf{27.6} \\ \hline
\end{tabular}
\caption{Ablation study of each proposed component in MemFusionMap evaluated on nuScenes with the new split at $50\,m$ range.}
\label{tab:ablation_each}
\vspace{-3mm}
\end{table}
\subsection{Ablation Studies}
We conduct extensive ablation studies to further break down the performance improvement brought by MemFusionMap and justify our design choice. We conduct all of the ablation studies using the new split of nuScenes at $100\times50\,m$ range. We include some ablation studies in the supplementary material.

\begin{table}[t] 
\centering
\begin{tabular}{l|c|l|c}
\hline
Module                                                                   & Memory Capacity $T_{\text{WM}}$ & \multicolumn{1}{c|}{mAP} & FPS \\ \hline
\multirow{5}{*}{\begin{tabular}[c]{@{}l@{}}Memory\\ Fusion\end{tabular}} & 1             & 24.3                     & \textbf{15.4}\\
                                                                         & 2             & 25.9                     & 15.2\\
                                                                         & 4             & \textbf{27.6}                     & 15.1\\
                                                                         & 6             & 26.2                     & 14.8\\
                                                                         & 8             & 27.0                       & 14.7\end{tabular}
\caption{Ablation study of the working memory capacity.}
\label{tab:ablation_memory_capacity}
\vspace{-6mm}
\end{table}

\begin{figure*}[t]
    \centering
    \includegraphics[width=0.98\linewidth]{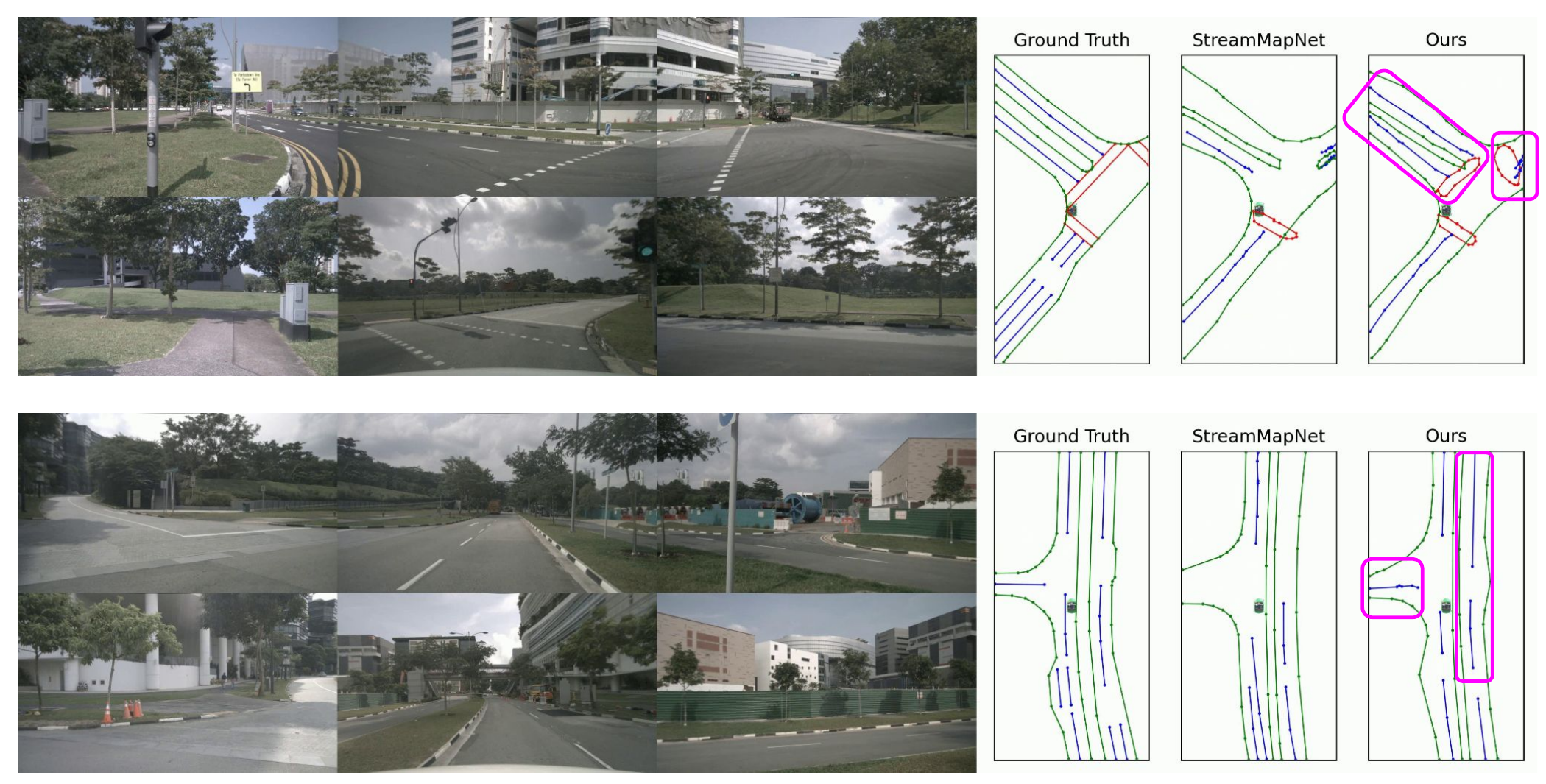}
    \caption{We show qualitative results on nuScenes~\cite{caesar2020nuscenes} at the extended $50\,m$ range. We show the input multi-view images on the left. In the HD maps, {\color{green}green} denotes road boundaries, {\color{red}red} denotes pedestrian crossings, and {\color{blue}blue} denotes lane lines. Best viewed in color.}
    \label{fig:qualitative}
    \vspace{-5mm}
\end{figure*}

We show an ablation study in \cref{tab:ablation_each}. We first remove the GRU-based temporal BEV fusion from StreamMapNet. We notice a notable performance drop, which shows the importance of temporal fusion for this task. We then test adding a convolutional working memory fusion module. It is observed that a simple convolution block can effectively improve temporal reasoning capability with the help of increased memory capacity by maintaining working memory features. The comparison between (a) and (c) also demonstrates the challenge of accumulating the entire history into a single memory latent. The improvement from (c) to (d) validates the importance of using dilated convolution for larger receptive field. Progressing from (d) to (e), the 1.7\% improvement on an already strong model justifies the significance of our novel design of the temporal overlap heatmap.

Furthermore, we study the effect of varying the working memory capacity $T_{\text{WM}}$. As shown in \cref{tab:ablation_memory_capacity}, we find that $T_{\text{WM}}=4$ is a sweet spot that delivers the best performance while also ensuring good runtime performance. We conjecture that the performance drop from further increasing $T_{\text{WM}}$ may be due to the unnecessary dilution of the current BEV feature when fusing with additional temporal information.

Additionally, we validate the design choice of dilation parameters of the convolution layers in the proposed working memory fusion module. Empirical results in \cref{tab:ablation_dilation} demonstrate that maintaining a symmetric dilation of $(2,2)$ brings the most improvement.

\begin{table}[t]
\centering
\begin{tabular}{l|c|l}
\hline
Module                                                                   & Conv Dilation & \multicolumn{1}{c}{mAP} \\ \hline
\multirow{4}{*}{\begin{tabular}[c]{@{}l@{}}Memory\\ Fusion\end{tabular}} & (1,1)         & 26.4                    \\
                                                                         & (2,2)         & \textbf{27.6}           \\
                                                                         & (2,3)         & 26.4                    \\
                                                                         & (3,3)         & 27.0                   
\end{tabular}
\caption{Ablation study of the dilation choice of convolution layers in the working memory fusion module. The conv dilation parameter corresponds to the BEV spatial dimension ($H$,$W$).}
\label{tab:ablation_dilation}
\vspace{-6mm}
\end{table}

\subsection{Qualitative Results}
We present qualitative comparison of MemFusionMap and StreamMapNet on nuScenes~\cite{caesar2020nuscenes} at the extended $100\times50\,m$ range, which is the most challenging configuration. We present several examples of MemFusionMap outperforming baseline model StreamMapNet~\cite{yuan2024streammapnet} in \cref{fig:qualitative}. Thanks to the proposed working memory fusion module with the temporal overlap heatmap, MemFusionMap predicts more accurate map elements with better geometry and fewer missed elements. We show more qualitative sequential examples in the supplementary material.

\section{Conclusion}
In this paper, we have proposed MemFusionMap, a novel approach for effective online vectorized HD map construction with enhanced temporal reasoning capability. We propose a working memory fusion module to improve the memory capacity of the model. We also design a temporal overlap heatmap to aid the temporal reasoning process across history frames. Our proposed MemFusionMap demonstrates state-of-the-art performance on open-source benchmarks with the geofenced splits and maintains a fast runtime and versatile design. We hope MemFusionMap will inspire future research works to leverage our proposed memory fusion module in other BEV perception tasks. We think integrating MemFusionMap in the MapTracker~\cite{chen2024maptracker} framework to leverage the tracking paradigm to further enhance the performance and temporal consistency is also worth exploring.

\section{Acknowledgement}
We deeply appreciate the fruitful discussion with Ziqi Pang. We also thank Xinyu Xie for the insights on leveraging working memory for enhancing temporal reasoning capabilities.

{\small
\bibliographystyle{ieee_fullname}
\bibliography{egbib}
}

\clearpage

\twocolumn[\centering
        \Large
        \textbf{MemFusionMap: Working Memory Fusion for Online\\Vectorized HD Map Construction}\\
        \vspace{0.5em}Supplementary Material \\
        \vspace{1.0em}]

\appendix

\section{Implementation Details}
In this section, we provide additional details of the MemFusionMap framework implementation.
We recognize the importance of improving reproducibility and will release code upon internal clearance and approval on the project website.

\subsection{Training}
We report additional details of training MemFusionMap. During training, we use the overall map loss $\mathcal{L}_{map}$ as defined in~\cite{yuan2024streammapnet}:
\begin{equation}
    \mathcal{L}_{map} = \lambda_1 \mathcal{L}_{Focal} + \lambda_2 \mathcal{L}_{line} + \lambda_3 \mathcal{L}_{trans},
\end{equation}
where $\mathcal{L}_{Focal}$ is the classification matching cost, $\mathcal{L}_{line}$ is the polyline-wise matching cost, $\mathcal{L}_{trans}$ is the auxiliary transformation loss for temporal query propagation, and $\lambda_1$-$\lambda_3$ are pre-defined loss weights. We set $\lambda_1$ as $5.0$, $\lambda_2$ as $50.0$, and $\lambda_3$ as $0.1$ for all of the experiment instances. In addition, as mentioned in the main paper, we are applying a two-stage training schedule following~\cite{park2022solofusion, yuan2024streammapnet}. To ensure fair comparison with the main baseline model (i.e. StreamMapNet~\cite{yuan2024streammapnet}), we follow the same schedule as described in the open-source repository of StreamMapNet. Specifically, as to the training sessions on nuScenes~\cite{caesar2020nuscenes} and Argoverse2~\cite{wilson2023argoverse}, the first single-frame training stage lasts $4$ and $5$ epochs, respectively. The second stage is a temporal training stage, where the input is consecutive sequences obtained from dividing the raw training sequence into two random sequences.

\subsection{Working Memory Fusion}
We provide further details about the implementation of the working memory fusion module. As mentioned in the main paper, we employ two convolution blocks to extract the temporal overlap heatmap feature ($\text{Conv}_{\mathbf{H}}$) and conduct working memory fusion to obtain the unified feature ($\text{Conv}_{\text{Mem}}$). 

As shown in Fig. 4 and Fig. 5 in the main paper, the temporal overlap heatmap includes low-level feature patterns that encode informative insights on vehicle trajectories. We apply $\text{Conv}_{\mathbf{H}}$ to extract the low level features from the heatmap $\mathbf{H}_t$:
\begin{equation}
\label{eq:conv_heatmap}
    \tilde{\mathbf{H}}_t = \text{sigmoid}(\text{Conv}_{\mathbf{H}}(\mathbf{H}_t)) \in \mathbb{R}^{C_{\mathbf{H}} \times H\times W}.
\end{equation}
We set $C_{\mathbf{H}}$ as 32, and report the convolution parameters in \cref{tab:conv_H_params}.

\begin{table}[h]
    \centering
    \begin{tabular}{|c|c|c|c|c|}
        \hline
        \textbf{Layer} & $C_{in}$ & $C_{out}$ & \textbf{Kernel} & \textbf{Activation} \\
        \hline
        1 & 1 & 16 & 3x3 & ReLU \\
        \hline
        2 & 16 & 16 & 3x3 & ReLU \\
        \hline
        3 & 16 & 32 & 1x1 & None \\
        \hline
    \end{tabular}
    \caption{Convolution block parameters in $\text{Conv}_{\mathbf{H}}$. We empirically find this configuration delivers stable performance with $\text{Conv}_{\text{Mem}}$.}
    \label{tab:conv_H_params}
    \vspace{-5mm}
\end{table}

In the working memory module, we fuse working memory features, $\mathcal{F}_\mathrm{WM}^{t}$, and the learned overlap heatmap, $\tilde{\mathbf{H}}_t$, with the encoded BEV feature, $\mathcal{F}_\text{BEV}^t$, via the convolutional memory fusion block, $\text{Conv}_{\text{Mem}}$, as follows:
\begin{equation}
    \tilde{\mathcal{F}}_\text{BEV}^t = \text{LayerNorm}(\text{Conv}_{\text{Mem}}(\text{Concat}(\mathcal{F}_\mathrm{WM}^{t}, \tilde{\mathbf{H}}_t, \mathcal{F}_\text{BEV}^t))),
\end{equation}
We show the details of the parameters of $\text{Conv}_{\text{Mem}}$ in \cref{tab:conv_mem_params}. The input channel dimension of the first layer is $1312$, which is computed as follows:
\begin{equation}
    C_{in} = T_{\text{WM}} \times C + C_{\mathbf{H}} + C,
\end{equation}
where $C$ is set to $256$, $T_{\text{WM}}$ is set to $4$ according to the ablation study in the main paper, and $C_{\mathbf{H}}$ is set to $32$. We study the most suitable dilation parameters and the working memory capacity in the ablation studies in the main paper.

\begin{table}[h]
    \centering
    \begin{tabular}{|c|c|c|c|c|c|}
        \hline
        \textbf{Layer} & $C_{in}$ & $C_{out}$ & \textbf{Kernel} & \textbf{Dilation} & \textbf{Activation} \\
        \hline
        1 & 1312& 256& 3x3 & (2,2) & ReLU \\
        \hline
        2 & 256& 256& 3x3 & (2,2) & ReLU \\
        \hline
        3 & 256& 256& 3x3 & (2,2) & ReLU \\
        \hline
    \end{tabular}
    \caption{Convolution block parameters in $\text{Conv}_{\text{Mem}}$.}
    \label{tab:conv_mem_params}
    \vspace{-6mm}
\end{table}

\section{Supplementary Experiments}
We provide additional experiments to further demonstrate the superior performance of MemFusionMap and justify our design choice.

\subsection{Faster Convergence}
\begin{figure}[h]
    \centering
    \includegraphics[width=1.0\linewidth]{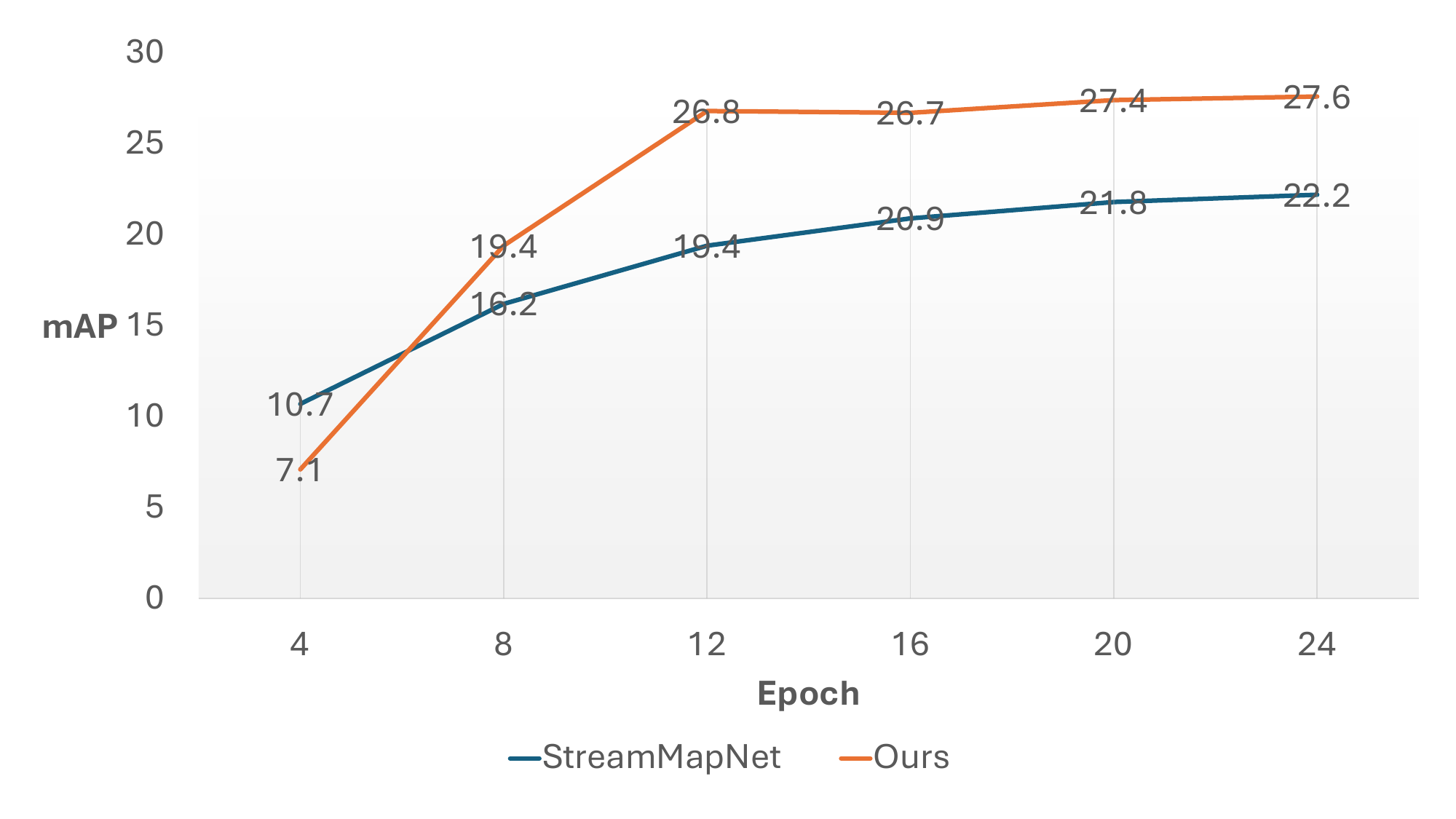}
    \caption{Convergence comparison based on experiments on the new nuScenes split at the $50\,m$ perception ranges.}
    \label{fig:convergence}
    \vspace{-4mm}
\end{figure}

When designing MemFusionMap, we prioritize scalability with the goal of integrating it into autonomous driving systems that utilize large-scale databases. As we expand to databases much larger than existing open-source benchmarks~\cite{caesar2020nuscenes, wilson2023argoverse}, it is crucial to maintain rapid convergence, which aims to obtain strong performance on the evaluation sets with a shorter training schedule. Motivated by this, we show a convergence plot of the mean Average Precision (mAP) score against the number of training epochs in \cref{fig:convergence}. We demonstrate MemFusionMap's capability of converging to a strong state with good performance using significantly fewer training epochs. We attribute this improvement to the improved temporal reasoning capability thanks to the innovative working memory fusion module and the use of a temporal overlap heatmap.

\subsection{Additional Ablation Studies}
We show additional ablation studies to validate our design choice for the temporal overlap heatmap. As shown in \cref{eq:conv_heatmap}, we use the sigmoid function as the non-linear activation function for the output of $\text{Conv}_{\mathbf{H}}$. As shown in \cref{tab:ablation_heatmap_activation}, using sigmoid delivers better mAP score. We conjecture the improvement from using sigmoid is that it can bound the output value to $[0,1]$, which works better to adapt to the test sets in which all sequences are full 20s, while the training sequences are split randomly into two sequences.

\begin{table}[t]
\centering
\begin{tabular}{l|c|l}
\hline
Module                                                                   & Non-linearity & \multicolumn{1}{c}{mAP} \\ \hline
\multirow{2}{*}{\begin{tabular}[c]{@{}l@{}}Temporal Overlap\\ Heatmap Feature\end{tabular}} & ReLU         & 26.5                    \\
                                                                         & Sigmoid         & \textbf{27.6}                             
\end{tabular}
\caption{Ablation study of the non-linear activation function choice for the module to extract the temporal overlap heatmap feature. The ablation study is conducted on the nuScenes~\cite{caesar2020nuscenes} using the new split~\cite{Roddick_2020_nusc_new_split} at the $100\times50\,m$ range.}
\label{tab:ablation_heatmap_activation}
\vspace{-6mm}
\end{table}

\begin{figure*}[t]
    \centering
    \includegraphics[width=0.98\linewidth]{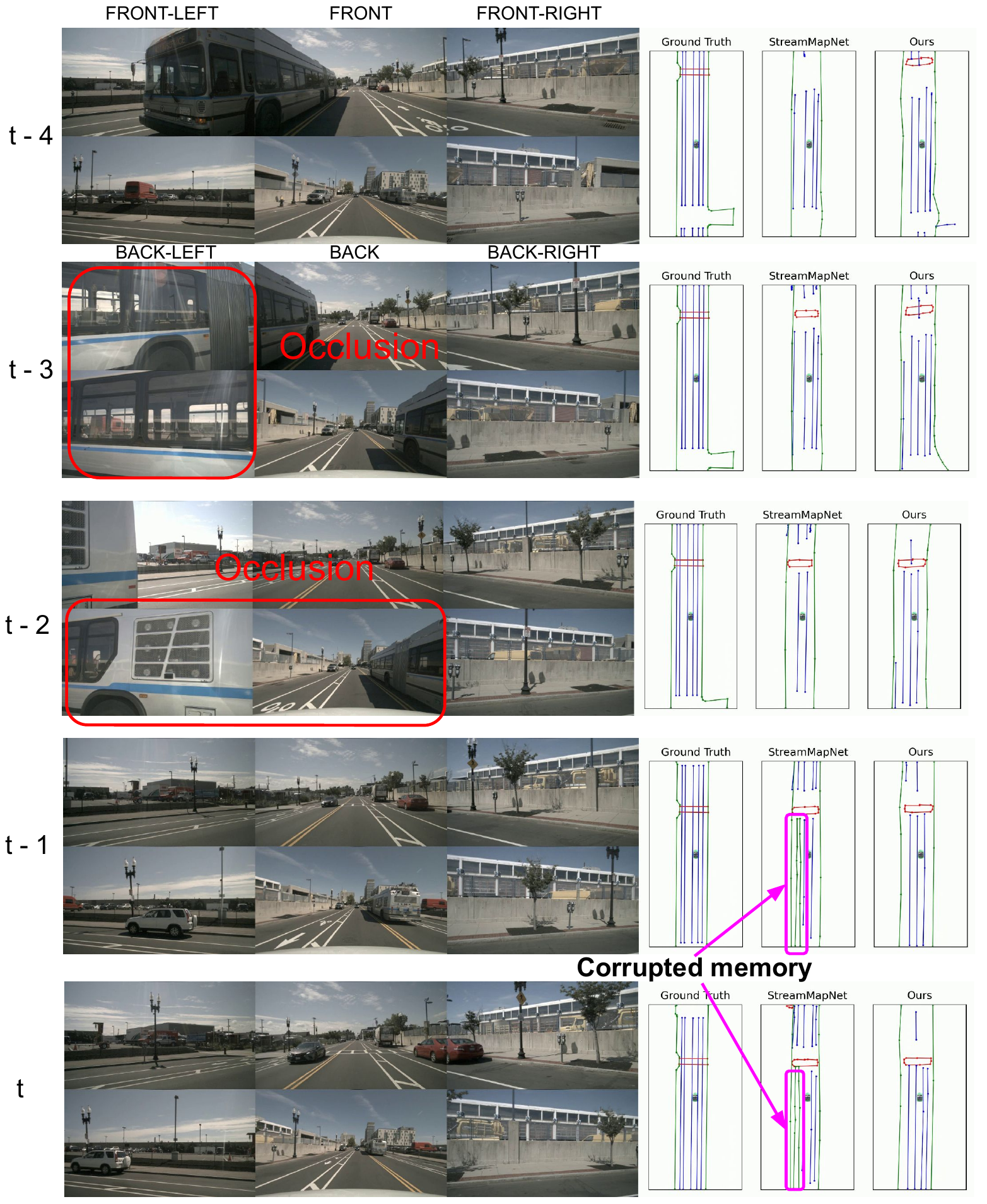}
    \caption{Qualitative results on nuScenes at the $100\times50\,m$ perception ranges. We show a consecutive sequence of $5$ frames. We show the input multi-view images on the left, with the camera placement in the first frame. In the HD maps, {\color{green}green} denotes road boundaries, {\color{red}red} denotes pedestrian crossings, and {\color{blue}blue} denotes lane lines. Best viewed zoomed in and in color.}
    \label{fig:supp_qualitative1}
\end{figure*}

\begin{figure*}[t]
    \centering
    \includegraphics[width=0.98\linewidth]{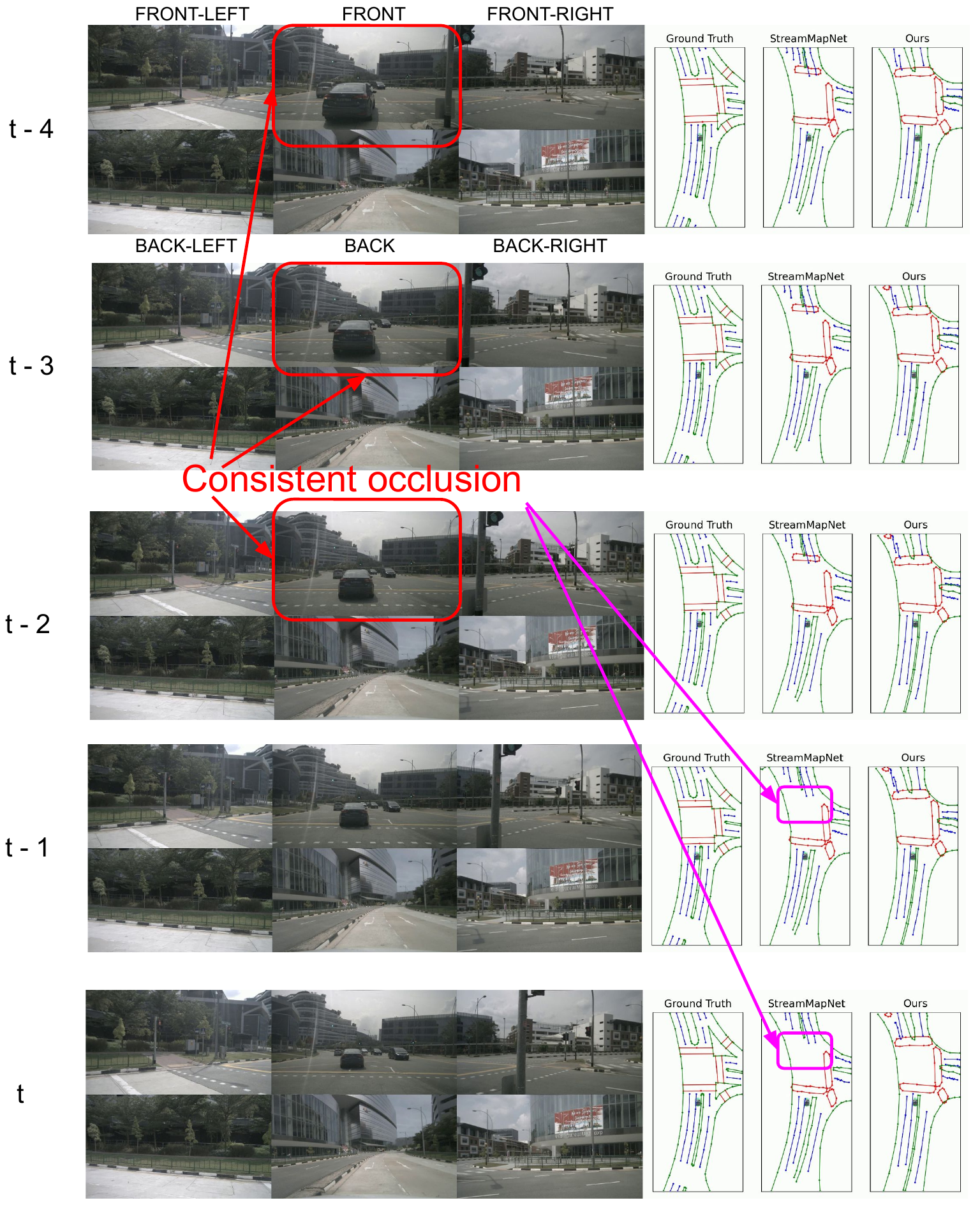}
    \caption{Qualitative results on nuScenes at the $100\times50\,m$ perception range. We show a consecutive sequence of $5$ frames. We show the input multi-view images on the left, with the camera placement in the first frame. In the HD maps, {\color{green}green} denotes road boundaries, {\color{red}red} denotes pedestrian crossings, and {\color{blue}blue} denotes lane lines. Best viewed zoomed in and in color.}
    \label{fig:supp_qualitative2}
\end{figure*}
\subsection{Qualitative Results}
As we have discussed in the main paper, the GRU-based~\cite{chung2014GRU} temporal fusion in StreamMapNet~\cite{yuan2024streammapnet} may struggle to effectively reason about the entire history due to limited memory capacity since the network only has access to the latest memory feature. \cref{fig:supp_qualitative1} and \cref{fig:supp_qualitative2} have demonstrated that MemFusionMap can notably mitigate this issue.

As shown in \cref{fig:supp_qualitative1}, StreamMapNet generates false predictions in frames $t-1$ and $t$, highlighted by {\color{magenta}magenta} boxes. It is surprising that the StreamMapNet model struggles with the highlighted region, given the good visibility from the input images at these frames. Upon back-tracing the history frames, we find that a bus quickly passes through the highlighted region, causing significant occlusion. Although false predictions do not occur in frames $t-3$ and $t-2$ due to temporal fusion, the challenging occlusion begins to introduce errors in the memory updates. As these errors accumulate, future frames (i.e., $t-1$ and $t$) suffer from corrupted memory features, leading to false predictions that could significantly harm downstream planning tasks. In contrast, MemFusionMap maintains a working memory fusion module and has access to features from the past $4$ frames, allowing it to access frame $t-4$, which provides good visibility of the highlighted region for frames $t-1$ and $t$.

A similar pattern is observed in \cref{fig:supp_qualitative2}. A pedestrian crossing missed by StreamMapNet is highlighted with {\color{magenta}magenta} boxes in frames $t-1$ and $t$. Examining frames from $t-4$ to $t$, we find that most of the highlighted region is consistently occluded by moving vehicles in the front camera, with only a small portion of it being intermittently visible. This complex scene with severe occlusion significantly challenges the model's temporal reasoning capability. This comparison validates the improved temporal reasoning capability of MemFusionMap, as it withstands the severe occlusion and consistently predicts the pedestrian crossing in the front, which is crucial to ensure safe navigation.

\section{Supplementary Discussion}
\subsection{Additional Future Work}
In MemFusionMap, the use of temporal overlap heatmap serves as an effective module to weigh the importance of historical BEV frames against the current frame. There remain interesting future works on the design of this module. For instance, there are no bounds on the value of the temporal overlap heatmap in MemFusionMap since each sequence in the evaluation datasets is clipped at 20 seconds, which does not usually hold in real-world deployment. Exploring how to maintain the temporal overlap heatmap to be properly bounded in a real-world long sequence is an interesting future work to improve the model's adaptability in real world. In addition, making the temporal overlap heatmap aware of occlusion is also worth exploring.

\subsection{Potential Negative Societal Impact}
Although MemFusionMap achieves significant improvements over existing models, it is important to note that it may still generate false predictions, particularly in complex scenarios and at long perception ranges, which could introduce safety concerns for autonomous driving systems. Additionally, while we specifically focus on geographically non-overlapping splits to prevent MemFusionMap from merely memorizing the training data and to ensure it generalizes to unseen environments, the model may still retain knowledge of the HD map data. This retention could raise further privacy and security issues.

\end{document}